\documentclass{article}

\usepackage[preprint]{neurips_2025}
\usepackage{caption}

\usepackage[utf8]{inputenc} 
\usepackage[T1]{fontenc}    
\usepackage{url}            
\usepackage{booktabs}       
\usepackage{amsfonts}       
\usepackage{nicefrac}       
\usepackage{microtype}      
\usepackage{xcolor}         
\usepackage{graphicx}
\usepackage{amsmath}
\newcommand{\method}{PlayerOne}
\usepackage{xspace}

\definecolor{citecolor}{HTML}{0071bc}
\definecolor{shadecolor}{rgb}{0.94,0.94,0.94}
\usepackage[colorlinks, linkcolor=red, colorlinks, anchorcolor=blue, citecolor=citecolor, pagebackref=True]{hyperref}

\title{PlayerOne: Egocentric World Simulator}

\author{
    Yuanpeng Tu$^{1,2}$ \thanks{Work during DAMO Academy internship. $\dagger$ Corresponding author.} \quad
    Hao Luo$^{2,3}$ \quad
    Xi Chen$^{1}$ \quad
    Xiang Bai$^{4}$ \quad
    Fan Wang$^{2}$ \quad
    Hengshuang Zhao$^{1,\dagger}$\\[2pt]
    $^{1}$HKU \quad
    $^{2}$DAMO Academy, Alibaba Group \quad
    $^{3}$Hupan Lab \quad
    $^{4}$HUST \\ [2pt]
    \textit{\href{https://playerone-hku.github.io/}{https://playerone.github.io}}
}
\begin{document}

\maketitle

\begin{figure}[ht]
\vspace{-15pt}
\begin{center}
	\includegraphics[width=1.0\linewidth]{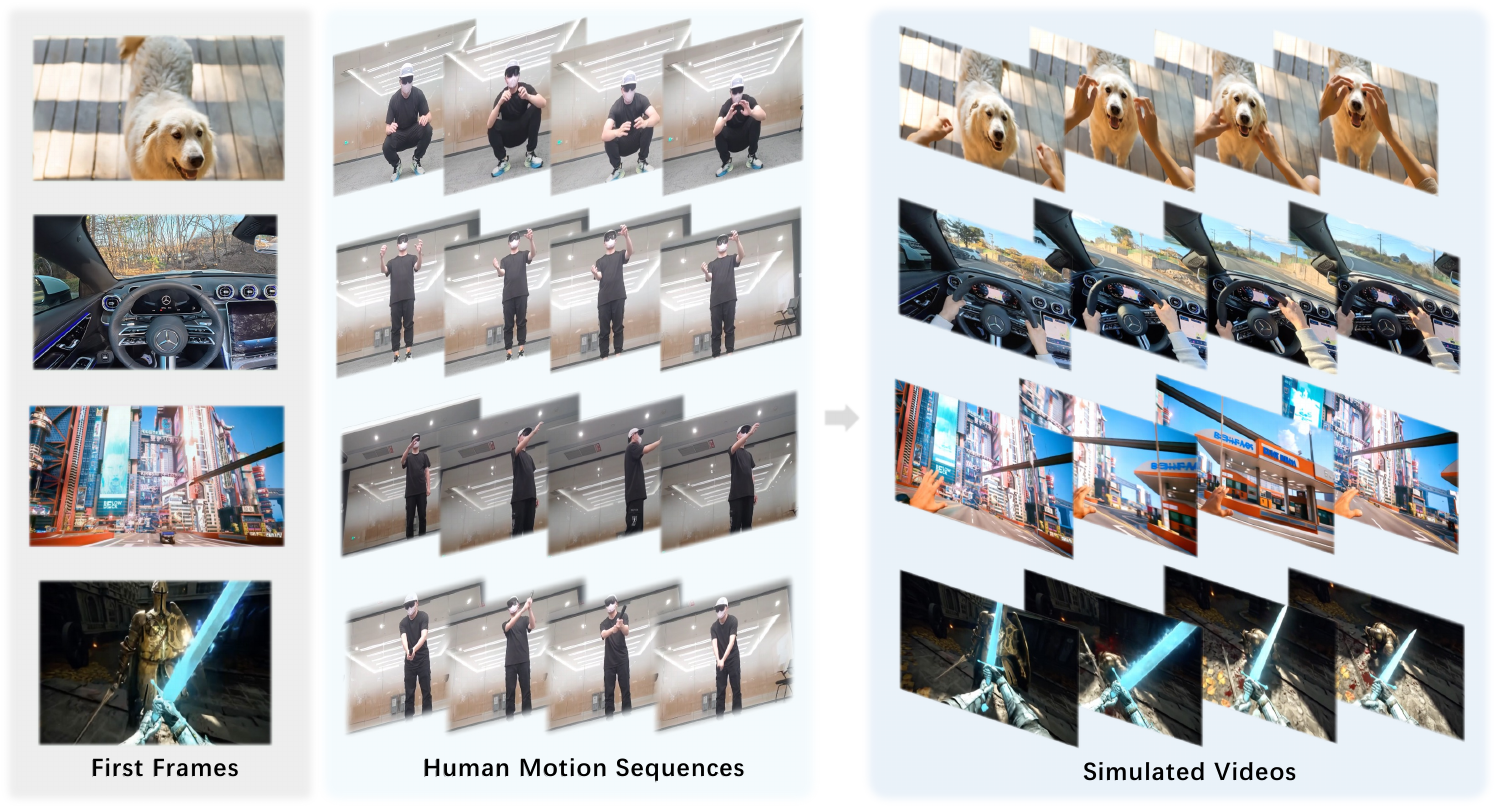}
\end{center}
\vspace{-10pt}
\caption{\small
\textbf{Simulated videos of our \method}. Given an egocentric image as the scene to be explored, we can simulate egocentric immersive videos that are accurately aligned with the user's motion sequence captured by an exocentric camera. All the users have been anonymized and action videos are shot with the front camera. 
}
\vspace{-2pt}
\label{fig:abs}

\end{figure}

\begin{abstract}

We introduce \method, the first egocentric realistic world simulator, facilitating immersive and unrestricted exploration within vividly dynamic environments. Given an egocentric scene image from the user, \method \xspace can accurately construct the corresponding world and generate egocentric videos that are strictly aligned with the real-scene human motion of the user captured by an exocentric camera. 
\method \xspace is trained in a coarse-to-fine pipeline that first performs pretraining on large-scale egocentric text-video pairs for coarse-level egocentric understanding, followed by finetuning on synchronous motion-video data extracted from egocentric-exocentric video datasets with our automatic construction pipeline.
Besides, considering the varying importance of different components, we design a part-disentangled motion injection scheme, enabling precise control of part-level movements. In addition, we devise a joint reconstruction framework that progressively models both the 4D scene and video frames, ensuring scene consistency in the long-form video generation. 
Experimental results demonstrate its great generalization ability in precise control of varying human movements and world-consistent modeling of diverse scenarios. It marks the first endeavor into egocentric real-world simulation and can pave the way for the community to delve into fresh frontiers of world modeling and its diverse applications.

\end{abstract}

\section{Introduction}
\label{sec:introduction}
World models~\cite{xiao2025worldmemlongtermconsistentworld,aether,wu2022daydreamer,hafner2019dreamer,nvidia2025cosmosworldfoundationmodel,muzero-general} have undergone extensive research due to their ability to model environmental dynamics and predict long-term outcomes. Recent breakthroughs in video diffusion models~\cite{wan2025,HaCohen2024LTXVideo,kong2024hunyuanvideo} have revolutionized this domain, enabling the synthesis of high-fidelity, action-conditioned simulations that forecast intricate future states. These advancements empower applications ranging from autonomous navigation in dynamic real-world environments to the creation of immersive, responsive virtual worlds in AAA game development. By bridging the gap between predictive modeling and interactive realism, world simulators are emerging as critical infrastructure for next-generation autonomous systems and game engines, particularly in scenarios requiring real-time adaptation to complex, evolving interactions.

Despite significant progress, this topic remains underexplored in existing research. Prior studies~\cite{guo2025mineworldrealtimeopensourceinteractive, xiao2025worldmemlongtermconsistentworld, feng2024matrixinfinitehorizonworldgeneration} predominantly focused on simulations within game-like environments, falling short of replicating realistic scenarios. Additionally, in their simulated environments, users are limited to performing predetermined actions (\textit{i.e.}, directional movements). Operating within the confines of a constructed world restricts the execution of unrestricted movements as in real-world scenarios. While some initial efforts~\cite{liang2024wonderlandnavigating3dscenes, ren2025gen3c3dinformedworldconsistentvideo, nvidia2025cosmosworldfoundationmodel} have been made toward real-world simulation, they mainly contribute to world-consistent generation without human movement control. Consequently, users are reduced to passive spectators within the environment, rather than being active participants. This limitation significantly impacts the user experience, as it prevents the establishment of a genuine connection between the user and the simulated environment.

Faced with these challenges, we aim to design an egocentric world foundational framework that enables the user being a freeform adventurer. Given a user-provided egocentric image as the world to be explored, it can enable the user to perform unrestricted human movements real-time captured by an exocentric camera and consistent 4D scene modeling in the simulated world. Specifically, we propose the first realistic egocentric world simulator termed \method. Starting from a diffusion transformer (DiT) model~\cite{Peebles2022DiT}, we first extract the latent of an egocentric user-input image. Meanwhile, we select the real-world human motions (\textit{i.e.}, human pose or keypoints) as our motion representation. Considering the varying importance of different body parts in our task, the human motion sequence is partitioned into three groups (\textit{i.e.}, head, hands, feet and body) and fed into our part-disentangled motion injection to generate latents that can enable precise part-wise control. Additionally, we developed a joint scene-frame reconstruction framework that can progressively complete scene point maps during the video generation process to enable scene-consistent generation. The DiT model takes the concatenation of the first frame latent, motion latent, video latent, and the point map latent as input and conducts noising and denoising on both the video and point map latent. Notably, the point map sequence is not required during inference, ensuring practical efficiency. Moreover, to overcome the absence of publicly available datasets, we curate required motion-video pairs from existing egocentric-exocentric datasets using an automated pipeline designed to filter and retain high-quality data. A coarse-to-fine training strategy is also designed to compensate for the data scarcity. The base model is fine-tuned on large-scale egocentric text-video data for coarse-level generation, then refined on our curated dataset to achieve precise motion control and scene modeling. Finally, we distill our trained model~\cite{yin2025causvid} to achieve real-time generation. By integrating these innovations, \method \xspace advances the field of dynamic world modeling. Our contributions are summarized as follows:

\begin{itemize}
\item We introduce \method, the first egocentric foundational simulator for realistic worlds, capable of generating video streams with precise control of highly free human motions and world consistency in real-time and exhibiting strong generalization in diverse scenarios.

\item We design a novel part-disentangled motion injection scheme to enhance fine-grained motion alignment, where a joint scene-frame reconstruction framework is introduced to guarantee world-consistent modeling in long-term video generation as well.

\item We construct an effective automatic dataset construction pipeline to extract high-quality motion-video pairs from existing egocentric-exocentric datasets, where a coarse-to-fine training scheme is also introduced to compensate for the data scarcity. 

\end{itemize}

\section{Related Work}
\label{sec:relatedwork}

\vspace{-2mm}

\noindent\textbf{Video generation.} The rapid development of diffusion models~\cite{rombach2021highresolution,ho2020denoising,zhang2023adding,latentdiffusion2022} has driven substantial advancements in video generation. Early researchers~\cite{guo2023animatediff,chen2023controlavideo} adapted existing text-to-image models to enable text-to-video generation to compensate for the limited availability of high-quality video-text datasets. Subsequently, diffusion transformers based frameworks~\cite{xie2024sana,Peebles2022DiT,kong2024hunyuanvideo,HaCohen2024LTXVideo,wan2025,genmo2024mochi} are proposed. When scaling-up training, they enable more highly realistic and temporally coherent generation results. Among them, HunyuanVideo~\cite{kong2024hunyuanvideo} substitutes T5 with a Multimodal Large Language Model. LTX-Video~\cite{HaCohen2024LTXVideo} modifies the VAE decoder to handle the final denoising step and convert latents into pixels. Wan~\cite{wan2025} introduces a full spatial-temporal attention to ensure computational efficiency.

\noindent\textbf{World models.} Existing world models~\cite{hafner2019dreamer,wu2022daydreamer,xiao2025worldmemlongtermconsistentworld,feng2024matrixinfinitehorizonworldgeneration,nvidia2025cosmosworldfoundationmodel,muzero-general} can be roughly divided into two categories: 1) Agent learning targeted models, 2) World simulation models. For the former, they~\cite{hafner2019dreamer,wu2022daydreamer,muzero-general} aim at enhancing policy learning within simulated environments. Among them, Dreamer~\cite{hafner2019dreamer} and DayDreamer~\cite{wu2022daydreamer} tackle long-horizon tasks by leveraging latent imagination. Meanwhile, MuZero~\cite{muzero-general} employs self-play Monte Carlo Tree Search (MCTS). In contrast, world simulation approaches focus on modeling environments by explicitly predicting next states given the current state and action. These methods prioritize enabling human-neural network interaction through high-fidelity rendering, robust control mechanisms. Recent advancements in video generation have further enabled the development of high-quality, controllable world simulations, sparking significant progress in the field of world simulation~\cite{guo2025mineworldrealtimeopensourceinteractive, ren2025gen3c3dinformedworldconsistentvideo,liang2024wonderlandnavigating3dscenes,xiao2025worldmemlongtermconsistentworld,feng2024matrixinfinitehorizonworldgeneration,nvidia2025cosmosworldfoundationmodel,aether,cheng2025animegamerinfiniteanimelife}. Among these works, WORLDMEM~\cite{xiao2025worldmemlongtermconsistentworld}. The Matrix~\cite{feng2024matrixinfinitehorizonworldgeneration} proposes the first world simulator capable of generating infinitely long real-scene video streams with real-time, responsive control. Matrix-Game~\cite{zhang2025matrixgame} redefines video generation as an interactive process of exploration and creation. Cosmos~\cite{nvidia2025cosmosworldfoundationmodel} presents a general-purpose world model and a pre-training-then-post-training scheme. Aether~\cite{aether} designs a unified framework with synergistic knowledge sharing across reconstruction, prediction, and planning objectives. However, these methods primarily focus on virtual game scenarios and are limited to specific directional actions, rather than facilitating high-degree-of-freedom motion control in real-world environments. To address these limitations, we target at developing a human motion driven realistic world simulator. Given an egocentric image, we can construct a real-scene world that immerses users as freeform adventurers with precise and unrestricted human motion control.

\vspace{-2mm}
\section{Method}
\label{sec:method}
\vspace{-2mm}

In this section, we detail the methodology of \method. Sec.~\ref{sec:overview} introduces the relevant preliminaries and the overall pipeline. Sec.~\ref{sec:components} presents the core of our proposed model, followed by Sec.~\ref{sec:strategy} introducing our dataset construction and training strategy.

\begin{figure*}[!t]
\begin{center}
	\includegraphics[width=1.0\linewidth]{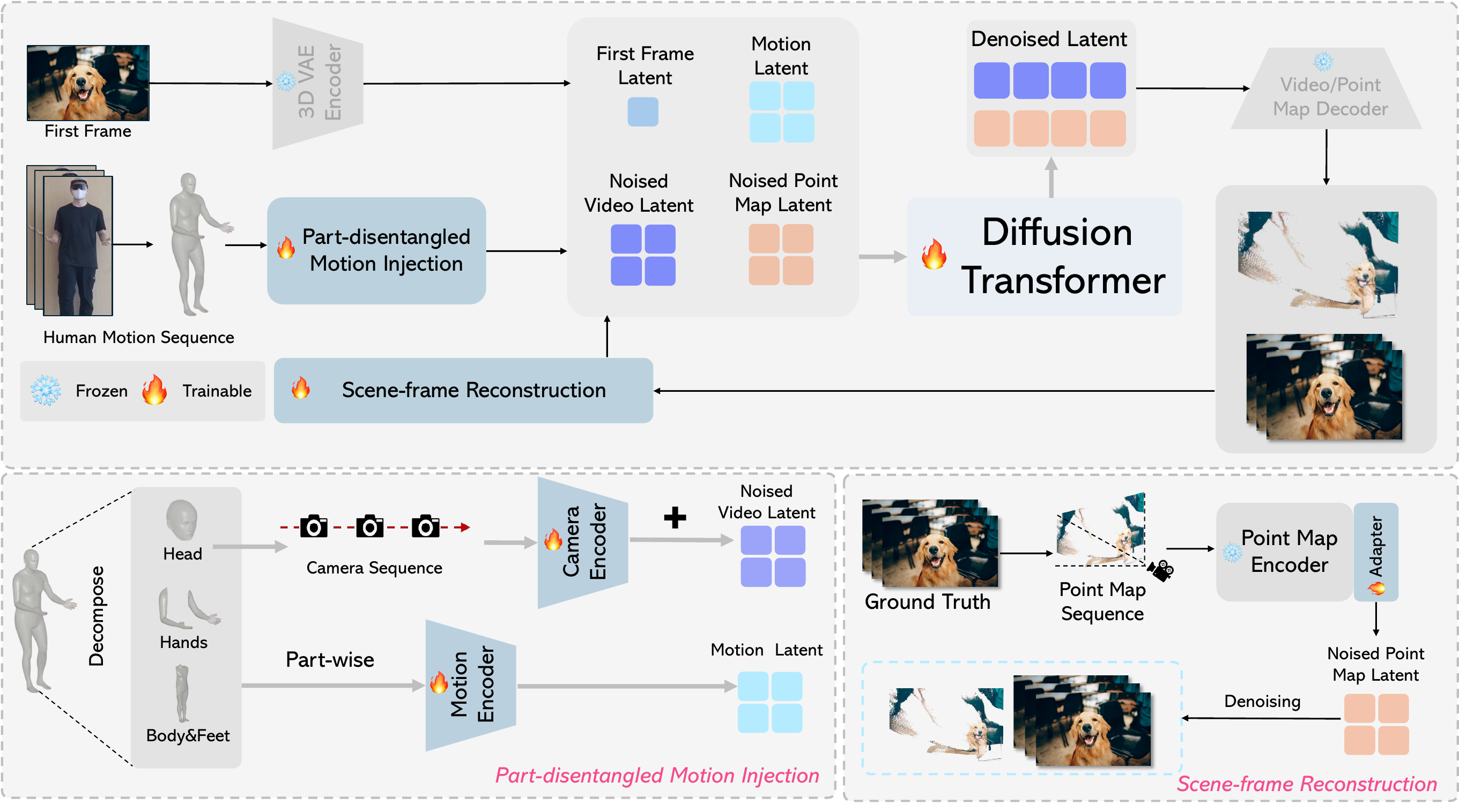}
\end{center}
\vspace{-10pt}
\caption{\small
\textbf{Overall framework of our \method}. It begins by converting the egocentric first frame into visual tokens. The human motion sequence is split into groups and fed into the motion encoders respectively to generate part-wise motion latents, with the head parameters converted into a rotation-only camera sequence. This camera sequence is then encoded via a camera encoder, and its output is injected into noised video latents to improve view-change alignment. Next, we render a 4D scene point map sequence with the ground truth video, which is then processed by a point map encoder with an adapter to produce scene latents. Then we input the concatenation of these latents into the DiT Model and perform noising and denoising on both the video and scene latents to ensure world-consistent generation. Finally, the denoised latents are decoded by VAE decoders to produce the final results. Note that only the first frame and the human motion sequence are needed for inference.
}
\vspace{-13pt}
\label{fig:framework}
\end{figure*}

\subsection{Overview}
\label{sec:overview}

Video diffusion models~\cite{Peebles2022DiT,ho2022videodiffusionmodels} consist of two key processes: a forward (noising) process and a reverse (denoising) process. The forward process gradually adds Gaussian noise, denoted as $\varepsilon \sim \mathcal{G}(0, \mathbf{I})$, to a clean latent sample $z_0 \in \mathbb{R}^{k \times c \times h \times w}$, where k, c, h, and w represent the dimensions of the video latents. This transforms $z_0$ into a noisy latent $z_t$. In the reverse process, a learned denoising model $\epsilon_\theta$ progressively removes the noise from $z_t$ to reconstruct the original latent representation.

As shown in Fig.~\ref{fig:framework}, our method comprises two core modules: Part-disentangled Motion Injection (PMI) and Scene-frame Reconstruction (SR). In PMI, we use real-scene human motion as the motion condition to enable free action control for the user. The first frame is converted into $z_{frame}$ via a 3D VAE encoder. The human motion sequence is split into three parts based on varying importance, and each part is fed into a 3D motion encoder to obtain latents. These motion latents are concatenated into $z_{motion} \in \mathcal{R}^{k \times 3 \times h \times w}$. To improve view alignment, we transform the head parameters of the human motion sequence into a camera sequence, which is then fed into a camera encoder. The output is added to the noised video latents $z_{video}$ to inject view-change signals. In SR, we jointly reconstruct video frames and 4D scenes to ensure world-consistent generation in the context of long video generation. We render a point map sequence from the ground truth video and feed it into a point map encoder with an adapter to obtain $z_{point} \in \mathcal{R}^{k \times 64 \times h \times w}$. Finally, all latents and conditions are concatenated channel-wise. The training objective of our method can be expressed as:

\begin{equation}
\mathcal{L}=\mathbb{E}_{}\left[\left\|\varepsilon-\varepsilon_\theta\left(\mathbf{z}_t, t\right)\right\|_2^2\right], \quad \text{where} \quad z_{t} = \sigma_{t}z_0+\beta_t\varepsilon, t \sim \mathcal{U}(0,1), \varepsilon \sim \mathcal{G}(0, \mathbf{I})
\end{equation}

Where $t = 1,\dots,T$, ${\sigma_t}^2 + {\beta_t}^2 = 1$. Since we only add noise to point map latents and video latents, thus $z_0 = z_{video}\otimes {z_{point}}$. $\otimes$ denotes the channel-wise concatenation operation, $\mathcal{U}(\cdot)$ represents a uniform distribution, and T denotes the denoising steps.

\vspace{-1mm}
\subsection{Model Components}
\label{sec:components}
\vspace{-1mm}
\noindent\textbf{Part-disentangled motion injection.} Prior studies~\cite{liang2024wonderlandnavigating3dscenes, ren2025gen3c3dinformedworldconsistentvideo,feng2024matrixinfinitehorizonworldgeneration,zhang2025matrixgame} typically utilize camera trajectories as motion conditions or are constrained to specific directional movements. These restrictions confine users to passive ``observer'' roles, preventing meaningful user interaction. In contrast, our approach empowers users to become active ``participants'' by adopting real-world human motion sequences (\textit{i.e.}, human pose or keypoints) as motion conditions, allowing for more natural and unrestricted movement.

However, our empirical analysis reveals that extracting latent representations holistically from human motion parameters complicates precise motion alignment. To address this challenge, we introduce a part-disentangled motion injection strategy that recognizes the distinct roles of various body parts. Specifically, hand movements are essential for interacting with objects in the environment, while the head plays a crucial role in maintaining egocentric perspective alignment. Accordingly, we categorize the human motion parameters into three groups: body and feet, hands, and head. Each group is processed through its own dedicated motion encoder, comprising eight layers of 3D convolutional networks, to extract the relevant latent features. This specialized processing ensures accurate and synchronized motion alignment. These latents are subsequently concatenated along the channel dimension to form the final part-aware motion latent representation $z_{motion} \in \mathcal{R}^{k \times 3 \times h \times w}$.

To further enhance the egocentric view alignment, we solely transform the head parameters of the human motion sequence into a sequence of camera extrinsics with only rotation values. We zero out the translation values in the camera extrinsics, assuming the head parameters are at the camera coordinate system's origin. Specifically, suppose the head parameter $\mathbf{v}=\left(\theta_x, \theta_y, \theta_z\right)$, we first normalize the rotation axis as follows:
\begin{equation}
\mathbf{u}=\frac{\mathbf{v}}{\|\mathbf{v}\|}, \quad \theta=\|\mathbf{v}\|
\end{equation}

Then we construct the rotation matrix as follows:
\begin{equation}
\mathbf{R}=\mathbf{I}+\sin \theta \cdot[\mathbf{u}]_{\times}+(1-\cos \theta) \cdot[\mathbf{u}]_{\times}^2
\end{equation}
Where $\mathbf{u}_{\times}$ is the cross product matrix of $\mathbf{u}$, which can be denoted as follows:
\begin{equation}
[\mathbf{u}]_{\times}=\left[\begin{array}{ccc}
0 & -u_z & u_y \\
u_z & 0 & -u_x \\
-u_y & u_x & 0
\end{array}\right]
\end{equation}

\begin{figure}[!t]
\vspace{-6pt}
\begin{center}
	\includegraphics[width=1.0\linewidth]{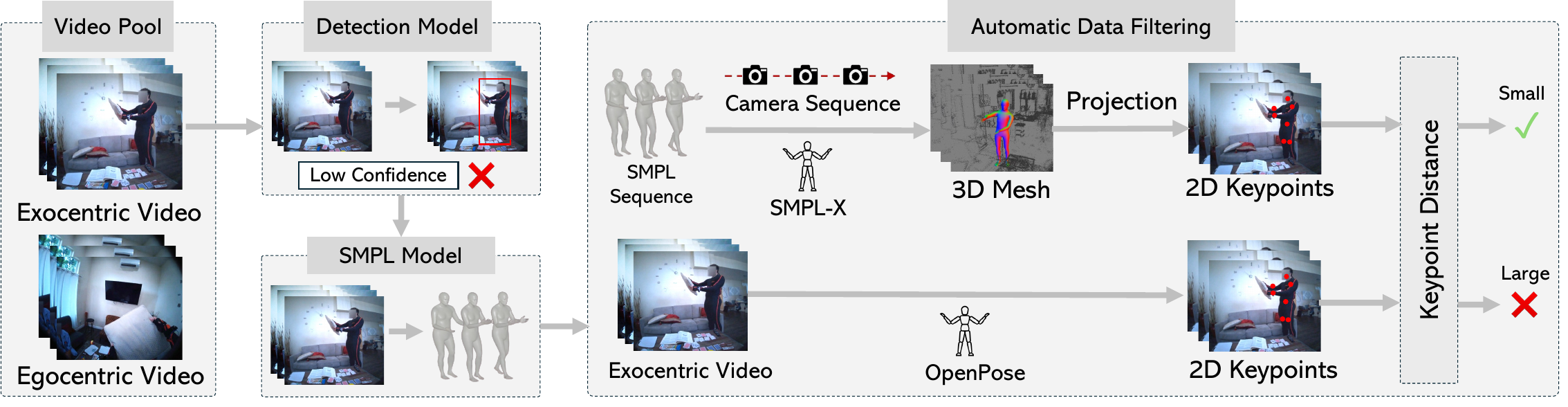}
\end{center}
\vspace{-10pt}
\caption{\small
\textbf{The overall pipeline of the dataset construction}. By seamlessly integrating detection and human pose estimation models, we can extract motion-video pairs from existing egocentric-exocentric video datasets while retaining high-quality data through our automatic filtering scheme. 
}
\vspace{-15pt}
\label{fig:pipeline}
\end{figure}

Then we use Plücker ray~\cite{zhang2024raydiffusion} to parameterize the camera extrinsics and then feed the output to an extra camera encoder, which shares a similar structure with the motion encoder. Then the latents from this encoder are added to the noised video latents to inject the view-change information.

\noindent\textbf{Scene-frame reconstruction.} While PMI enables precise control over egocentric perspective and motion, it does not guarantee scene consistency within the generated world. To address this limitation, we introduce a joint reconstruction framework that simultaneously models the 4D scene and video frames, ensuring scene coherence and continuity throughout the video. Specifically, it begins by employing CUT3R~\cite{wang2025continuous} to generate a point map for each frame based on ground truth video data, reconstructing the n-th frame’s point map using information from frames 1 through n. These point maps are then compressed into latent representations using a specialized point map encoder~\cite{Geo4D}. To integrate these latents with video features, we implement an adapter composed of five 3D convolutional layers. This adapter aligns the point map latents with video latents and projects them into a shared latent space, facilitating seamless integration of motion and environmental data. Finally, we concatenate the latent representations from the first frame, the human motion sequence, the noised video latents, and corresponding noised point map latents. This comprehensive input is then fed into a diffusion transformer for denoising, resulting in a coherent and visually consistent world. Importantly, point maps are only required during the training phase. \textit{During inference, the system simplifies the process by utilizing only the first frame and the corresponding human motion sequence to generate world-consistent videos.} This streamlined approach enhances generation efficiency while ensuring that the resulting environment remains stable and realistic throughout the entire video.

\vspace{-1mm}

\subsection{Training Strategy}
\label{sec:strategy}
\vspace{-1mm}

\noindent\textbf{Dataset preparation.} The ideal training samples for our task are egocentric videos paired with corresponding motion sequences. However, no such dataset currently exists in publicly available repositories. As a substitute, we derive these data pairs from existing egocentric-exocentric video datasets through an automatic pipeline. Specifically, for each synchronized egocentric-exocentric video pair, we first employ SAM2~\cite{ravi2024sam2} to detect the largest person in the exocentric view. The background-removed exocentric video is then processed using SMPLest-X~\cite{yin2025smplest} to extract the SMPL parameters of the identified individual as the human motion. To enhance optimization stability, an L2 regularization prior is incorporated. We then evaluate the 2D reprojection consistency to filter out low-quality SMPL data. This involves generating a 3D mesh from the SMPL parameters using SMPLX~\cite{SMPLX}, projecting the 3D joints onto the 2D image plane with the corresponding camera parameters, and extracting 2D key points via OpenPose~\cite{cao2019openpose}. The reprojection error is calculated by measuring the distance between the SMPL-projected 2D key points and those detected by OpenPose. Data pairs with reprojection errors in the top 10\% are excluded, ensuring a final dataset of high-quality motion-video pairs. The refined SMPL parameters are decomposed into body and feet (66 dimensions), head orientation (3 dimensions), and hand articulation (45 dimensions per hand) components for each frame. These components are fed into their respective motion encoders. The dataset construction pipeline is illustrated in Fig.~\ref{fig:pipeline}. As detailed in Tab.~\ref{tab:dataset}, our training dataset combines multiple publicly available datasets to ensure comprehensive coverage of diverse environmental contexts, action types, and intensity levels, thereby enhancing model generalization.

\begin{table}[t]
\caption{%
    \textbf{Statistics of datasets} used for training our \method. ``quality'' particularly refer to the image resolution. ``Ego-Exo'' denotes whether the dataset contains egocentric-exocentric video pairs. 
}
\label{tab:dataset}
\centering
\resizebox{1.0\linewidth}{!}{
\setlength{\tabcolsep}{14pt}
\begin{tabular}{lcccccccc}
\toprule

Dataset &  EgoExo-4D~\cite{grauman2024ego}           &  Nymeria~\cite{ma24eccv}                         &  FT-HID~\cite{Guo2022FT-HID}    & EgoExo-Fitness~\cite{li2024egoexo}  & Egovid-5M~\cite{wang2024egovid}                   \\
\midrule

Size            &       740          &   1,200               &         38,364          & 1,276   &    5M       \\
Resolution            &  1080p               &       1408p            &    1080p            &     1080p        &  1080p  \\
Ego-Exo            &        $\checkmark$               & $\checkmark$      &      $\checkmark$          &   $\checkmark$        &    $\times$  \\
\bottomrule
\end{tabular}}
\vspace{-4mm}

\end{table}

\noindent\textbf{Coarse-to-fine training.} Though we can extract high-quality motion-video training data with our automatic pipeline, the limited scale of this dataset is insufficient for training video generation models to produce high-quality egocentric videos. To address this, we harness the extensive egocentric text-video datasets (\textit{i.e.}, Egovid-5M~\cite{wang2024egovid}). Specifically, we first fine-tune the baseline model using LoRA on large-scale egocentric text-video data pairs, enabling egocentric video generation with coarse-level motion alignment. Then we freeze the trained LoRA and fine-tune the last six blocks of the model with our constructed high-quality dataset to enhance fine-grained human motion alignment and view-invariant scene modeling, which can effectively address the scarcity of pair-wise data. Finally, we adopt an asymmetric distillation strategy that supervises a causal student model with a bidirectional teacher~\cite{yin2025causvid} to achieve real-time generation and long-duration video synthesis.

\vspace{-3mm}
\section{Experiments}
\label{sec:experiments}
\vspace{-2mm}

\subsection{Experimental Setting}
\vspace{-1mm}
\noindent\textbf{Implementation details.} We choose Wanx2.1 1.3B~\cite{wan2025} as the base generator. We set the LoRA rank and the update weight of the matrices as 128 and 4 respectively and initialize its weight following ~\cite{wan2025}. The inference step and the learning rate are set as 50 and $1\times10^{-5}$ respectively, where the Adam optimizer and mixed-precision bf16 are adopted. The cfg of 7.5 is used. We train our model for 100,000 steps on 8 NVIDIA A100 GPUs with a batch size of 56 and sample resolution of 480$\times$480. The generated video runs at eight frames per second, and we utilize 49 video frames (6 seconds) for training. After distillation, our method can achieve 8 FPS to generate the desired results. \textit{All the action videos in this paper are shot with the front camera.}

\begin{figure*}[!t]
\begin{center}
	\includegraphics[width=1.0\linewidth]{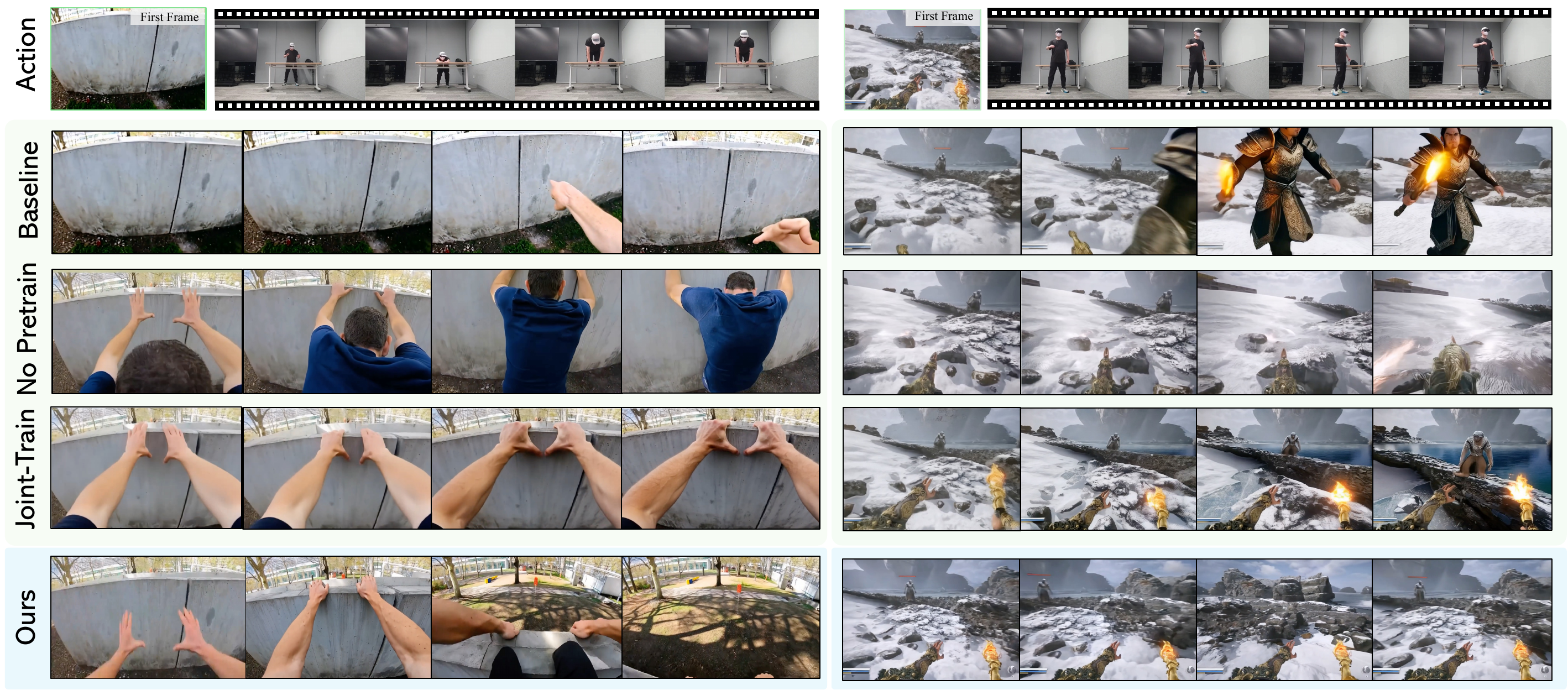}
\end{center}
\vspace{-9pt}
\caption{\small
\textbf{Investigation on coarse-to-fine training}. ``Joint-Train'' and ``No Pretrain'' denote training with both motion-video pairs and large-scale egocentric videos in a one-stage manner and training with only motion-video pairs respectively. The Wanx2.1 1.3B is adopted as the baseline.
}
\vspace{-16pt}
\label{fig:traingscheme}
\end{figure*}

\vspace{-0.5mm}

\noindent\textbf{Benchmark.} Since there is no publicly available benchmark for our task, we construct a benchmark with 100 videos collected from Nymeria~\cite{ma2024nymeria} dataset, which is not included for training. It consists of coarse-level motion descriptions for each sample and covers diverse realistic scenarios. Considering the information gap between the human motion sequence and the text, we further use Qwen2.5-VL~\cite{qwen} to enrich the caption to generate videos for the competitors for more fair comparisons.
\vspace{-0.5mm}

\noindent\textbf{Metrics.} On our constructed benchmark, for evaluation of alignment with the given text descriptions, we calculate both CLIP-Score and DINO-Score, where SSIM, and LPIPS~\cite{zhang2018unreasonableeffectivenessdeepfeatures} are employed to evaluate the video fidelity of the generated video. Besides, we calculate the frame consistency to evaluate the temporal coherence and consistency of the generated video frames over time. We further utilize a 3D hand pose estimation model~\cite{prakash20243dhandposeestimation} to estimate the hand pose of the generated videos and use the results of the ground truth video as the labels. Afterward, we follow ~\cite{prakash20243dhandposeestimation} to calculate two metrics: (1) Mean Per-Joint Position Error (MPJPE): the L2 distance between the predicted and ground truth joints for each hand after subtracting the root joint. (2) Mean Relative-Root Position Error (MRRPE): the metric distance between the root joints of the left hand and right hand.

\vspace{-2mm}
\subsection{Ablation Study}
\vspace{-2mm}

\noindent\textbf{Investigation on coarse-to-fine training.} We first evaluate several variants of our coarse-to-fine training scheme, as depicted in Fig.~\ref{fig:traingscheme}. Specifically, when inputting action descriptions into the baseline model without fine-tuning, the generated results exhibit noticeable flaws. Similar issues can be observed when training with only motion-video pairs. We also explore jointly training with both large-scale egocentric videos and motion-video pairs. Specifically, when inputting egocentric videos, we set the motion latent values to zero and extract the latents of the text description to serve as the motion condition. Despite this variant being capable of generating egocentric videos, it fails to produce results accurately aligned with the given human motion conditions. In contrast, our coarse-to-fine training scheme delivers much better outcomes compared to these variants.

\begin{figure*}[!t]
\begin{center}
	\includegraphics[width=1.0\linewidth]{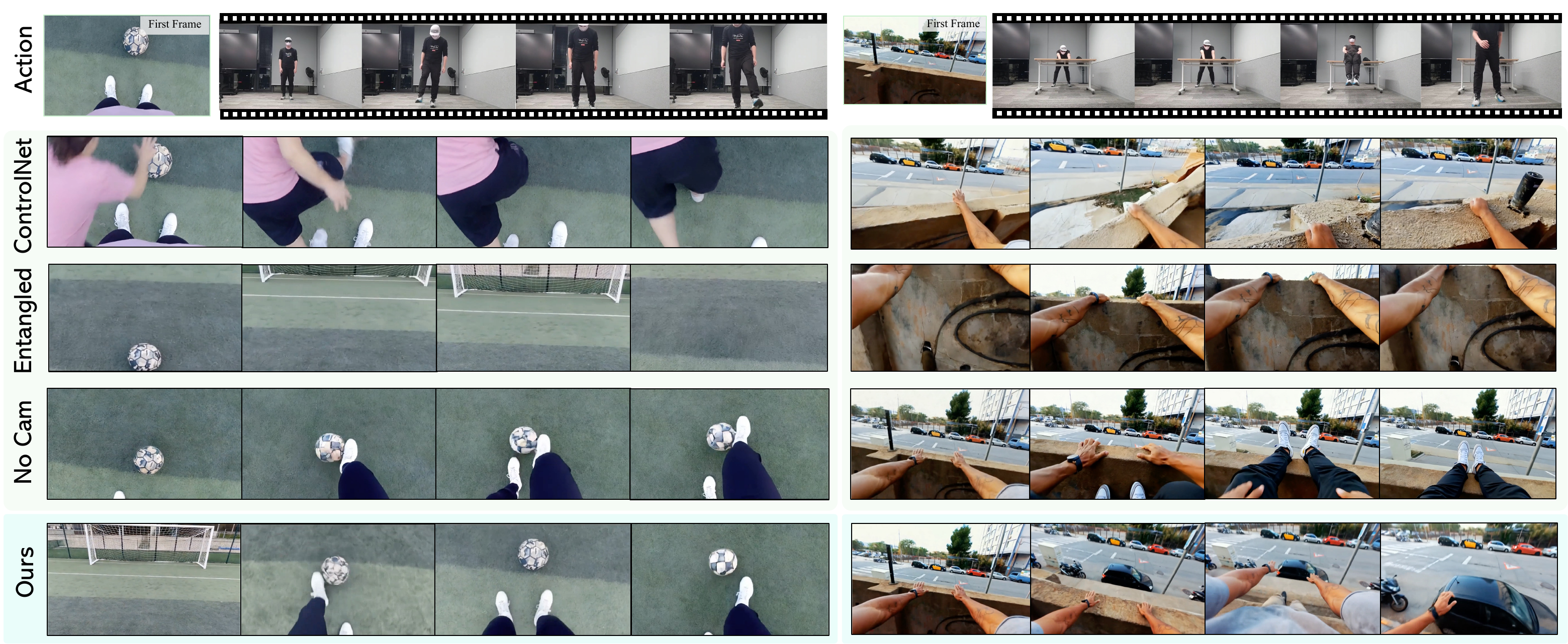}
\end{center}
\vspace{-10pt}
\caption{\small
\textbf{Investigation on part-disentangled motion injection}. ``ControlNet'' denotes injecting motion latents with a ControlNet~\cite{zhang2023adding}. ``Entangled'' and ``No Cam'' denote inputting the whole motion sequence into a motion encoder without dividing into groups and removing the camera encoder respectively.
}
\vspace{-10pt}
\label{fig:pmi}
\end{figure*}

\vspace{-1mm}

\noindent\textbf{Investigation on part-disentangled motion injection.} Next, we conduct a detailed analysis of our PMI module. Specifically, three variants are included: ControlNet-based~\cite{zhang2023adding} motion injection, inputting motion sequences as a unified entity (the ``Entangled'' scheme), and removing our camera encoder. As shown in Fig.~\ref{fig:pmi}, the ControlNet-based scheme suffers from information loss, preventing it from producing results that accurately align with the specified motion conditions. Similarly, the entangled scheme demonstrates comparable shortcomings. Furthermore, removing the camera encoder leads to the model's inability to generate view-accurate alignments. As depicted in Fig.~\ref{fig:pmi}, this variant fails to produce the corresponding perspective change associated with crouching. Ultimately, our PMI module successfully generates outcomes that are both view-aligned and action-aligned.

\begin{figure*}[!t]
\begin{center}
	\includegraphics[width=1.0\linewidth]{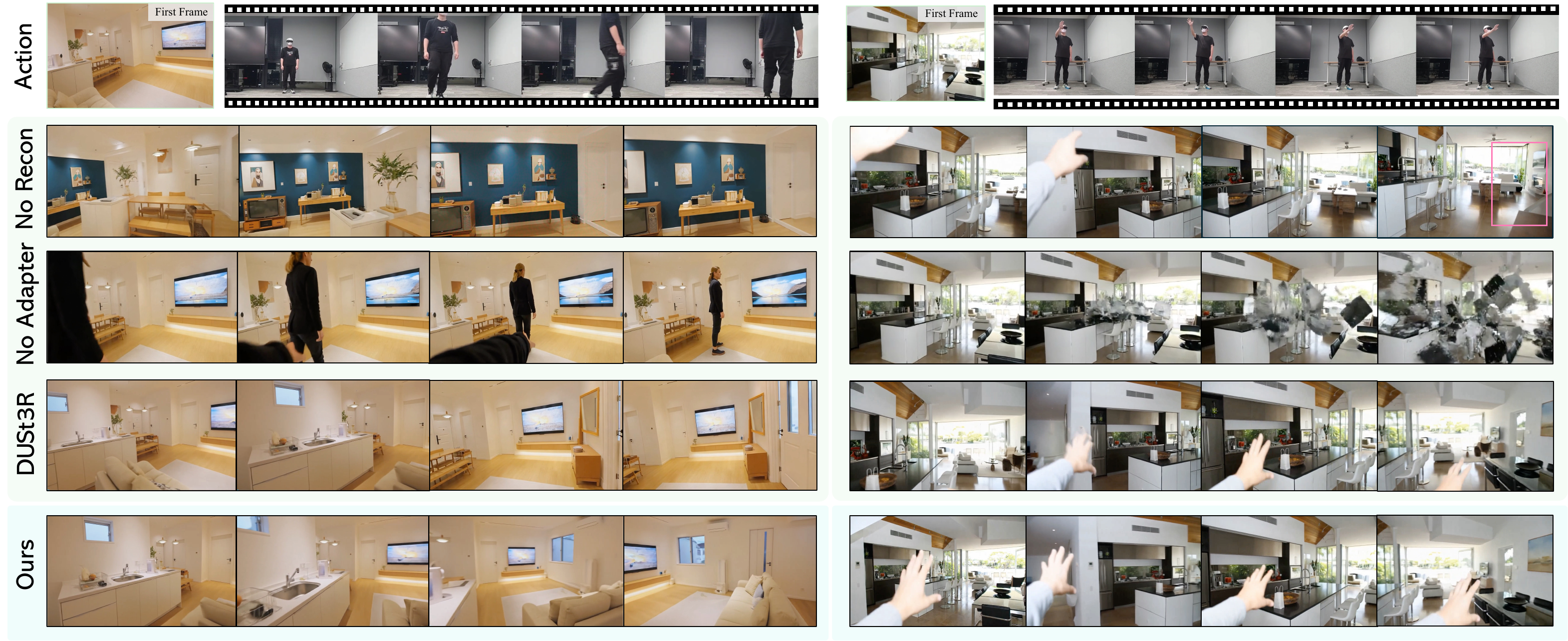}
\end{center}
\vspace{-9pt}
\caption{\small
\textbf{Investigation on scene-frame reconstruction}. ``No Recon''/``No Adapter'' denote training without reconstruction/the adapter. ``DUStR'' is replacing CUT3R with DUStR for point map rendering. 
}
\vspace{-8pt}
\label{fig:sr}
\end{figure*}

\begin{figure*}[!t]
\vspace{-6pt}
\begin{center}
	\includegraphics[width=1.0\linewidth]{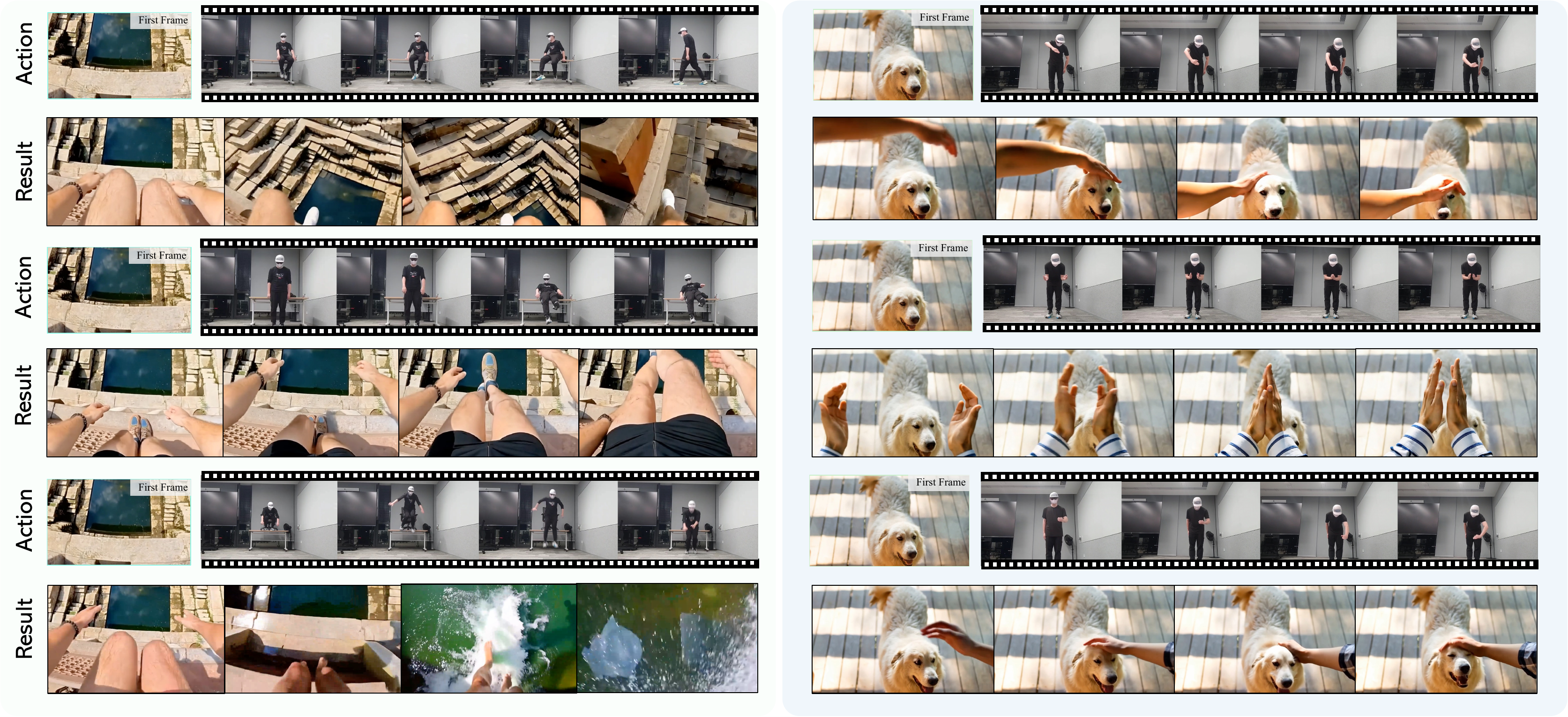}
\end{center}
\vspace{-9pt}
\caption{\small
\textbf{Qualitative evaluation on the motion alignment}. We generate simulated videos based on the same first frame but different motion sequences. Results show that we can achieve accurate motion alignment.
}
\vspace{-14pt}
\label{fig:ablationstudy_smpl}
\end{figure*}

\begin{table*}[!t]
\caption{%
    \textbf{Quantitative evaluation} on the components of \method. \method \xspace outperforms all these variants. ``No Camera''/``Filtering'' denote training without/with the camera encoder/data filtering.
}
\vspace{-2mm}
\label{tab:automatic_ablation}
\centering
\renewcommand{\arraystretch}{1.2}
\resizebox{1.0\linewidth}{!}{
\setlength{\tabcolsep}{1pt}
\begin{tabular}{lccccccc}
\toprule

& DINO-Score~($\uparrow$)   & CLIP-Score~($\uparrow$)  & MPJPE~($\downarrow$) & MRRPE~($\downarrow$)    & FVD~($\downarrow$) & LPIPS($\downarrow$)  \\
\midrule
Baseline                                      &51.3&65.6&376.14&341.01&394.16&0.1421 \\
+ Pretrain                                    &56.6&74.4&258.05&232.17&301.32&0.1146 \\
\hline
+ Pretrain\&ControlNet                        &57.1&75.2&241.73&218.46&287.52&0.1103 \\
+ Pretrain\&Entangled                         &58.0&76.3&235.12&212.53&279.41&0.1060  \\
+ Pretrain\&PMI (No Camera)                   &60.7&79.8&183.25&196.35&257.04&0.0902  \\
+ Pretrain\&PMI                               &62.5&81.3&156.76&175.18&245.72&0.0839  \\
\hline
+ Pretrain\&PMI\&Filtering                    &64.2&83.8&141.56&163.04&230.50&0.0782 \\
+ Pretrain\&PMI\&Filtering\&Recon(No Adapter) &62.7&81.6&176.23&180.10&240.17&0.0919  \\
+ Pretrain\&PMI\&Filtering\&Recon(DUSt3R)     &67.5&87.7&129.08&152.22&228.20&0.0685  \\
\hline
\method (ours)  & \textbf{67.8}                &\textbf{88.2}                  & \textbf{127.16}   &\textbf{151.62}&\textbf{226.12}&\textbf{0.0663} \\
\bottomrule
\end{tabular}}
\vspace{-6mm}

\end{table*}

\vspace{-1mm}

\noindent\textbf{Investigation on scene-frame reconstruction.} Additionally, we conducted a detailed analysis of the SR module, exploring three variants: omitting reconstruction, removing the adapter within the SR module, and substituting CUT3R~\cite{wang2025continuous} with DUStR~\cite{Wang_2024_CVPR} for point map rendering. As illustrated in Fig.~\ref{fig:sr}, the absence of reconstruction results in the model's inability to generate consistently simulated results. Moreover, training without the adapter leads to noticeable distortions. Furthermore, after replacing CUT3R~\cite{wang2025continuous} with DUStR~\cite{Wang_2024_CVPR}, our \method \xspace can also produce scene-consistent outputs, demonstrating its robustness to different point map rendering techniques.

\vspace{-0.5mm}

\noindent\textbf{Motion alignment.} To verify the alignment capability with the given motion condition, we conduct experiments by generating world-simulated videos with the same first frame but different human motion sequences.  Fig.~\ref{fig:ablationstudy_smpl} shows that our \method \xspace can accurately generate corresponding results according to different conditions and produce reasonable interactive changes. 
\vspace{-0.5mm}

\noindent\textbf{Quantitative comparisons.} We provide quantitative results on the core components of our \method \xspace in Tab.~\ref{tab:automatic_ablation}. All numerical results concur with the visualization outcomes. A significant performance improvement is observed when the model undergoes pre-training on large-scale egocentric text-video datasets. The introduction of PMI yields an additional accuracy boost, and it outperforms all of its variants. In addition, our designed filtering strategy maximizes performance as well by filtering noisy motion-video pairs. By introducing our joint scene-frame reconstruction scheme, we achieve superior results across all metrics. 
\vspace{-0.5mm}

\vspace{-1.5mm}
\subsection{Comparison with State-of-the-arts}
\vspace{-1.5mm}

\begin{figure*}[!t]

\begin{center}
	\includegraphics[width=1.0\linewidth]{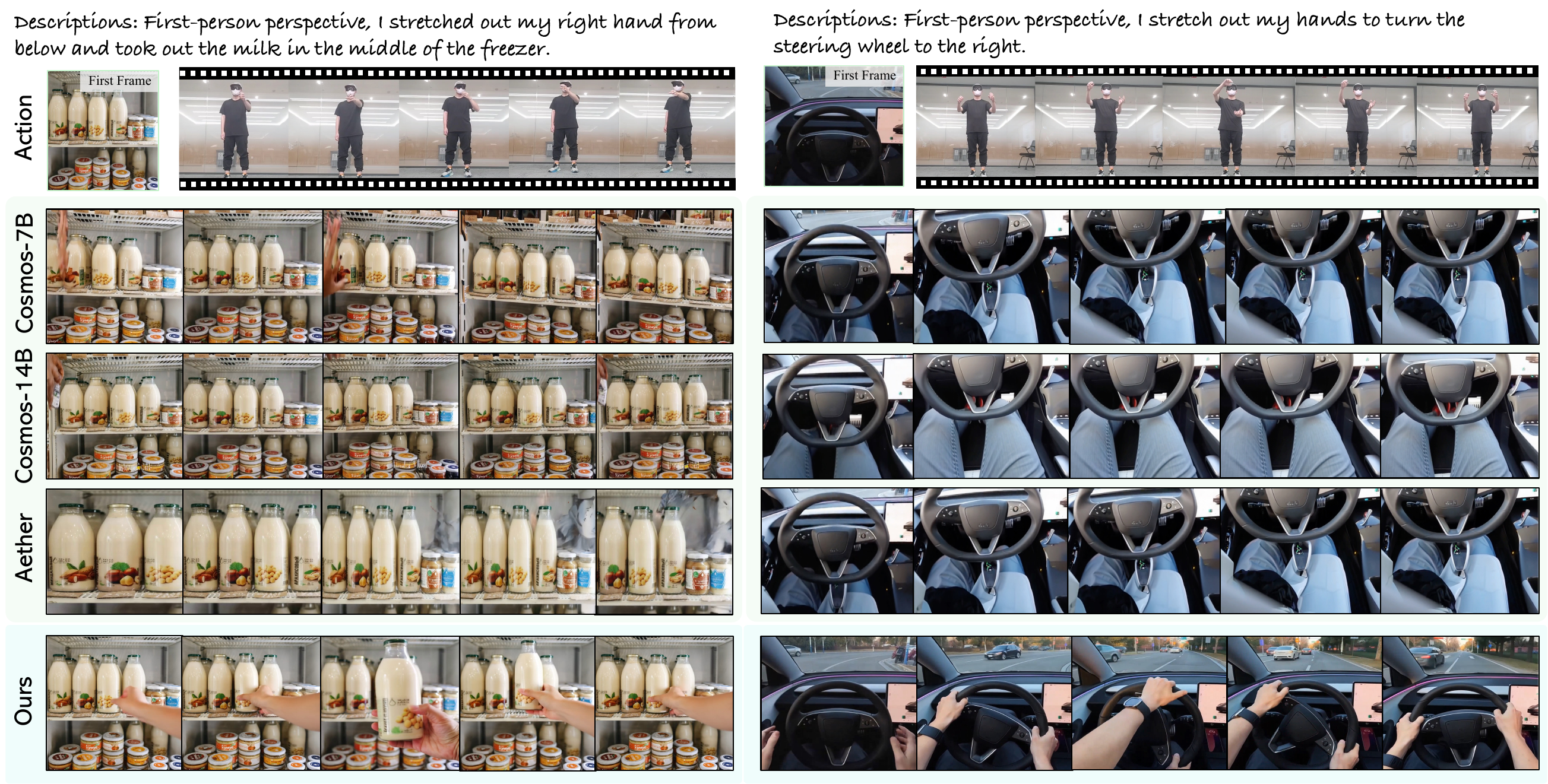}
\end{center}
\vspace{-9pt}
\caption{\small
\textbf{Qualitative comparisons between our method and other competitors}. Our \method \xspace can achieve the best performance on both the motion alignment, video quality. 
}
\vspace{-9pt}
\label{fig:qualitative}
\end{figure*}

\begin{table*}[!t]
\caption{%
    \textbf{Quantitative comparison} between our \method \xspace and other works. Seven metrics are employed for the evaluation. \method \xspace outperforms these methods across all the metrics.
}
\vspace{-2mm}

\label{tab:automatic_psnr}
\centering
\resizebox{1.0\linewidth}{!}{
\setlength{\tabcolsep}{6pt}
\begin{tabular}{lccccccc}
\toprule
& DINO-Score~($\uparrow$)   & CLIP-Score~($\uparrow$)  & MPJPE~($\downarrow$) & MRRPE~($\downarrow$)    & FVD~($\downarrow$) & LPIPS($\downarrow$)  \\
\midrule
Aether~\cite{aether}                                                  &38.0&64.2&415.70&431.05&397.40& 0.1856               \\
Cosmos(Diff-7B)~\cite{nvidia2025cosmosworldfoundationmodel}           &45.3&70.3&301.92&324.12&346.09 & 0.1630           \\
Cosmos(Diff-14B)~\cite{nvidia2025cosmosworldfoundationmodel}          &51.6&79.7&256.73&253.06&302.17&  0.1351             \\

\method (ours)  & \textbf{67.8}                &\textbf{88.2}                  & \textbf{127.16}   &\textbf{151.62}&\textbf{226.12}&\textbf{0.0663}               \\
\bottomrule
\end{tabular}}
\vspace{-3mm}

\end{table*}

\begin{table*}[!t]
\caption{
    \textbf{User study} on our \method \xspace and existing alternatives. “Quality”, “Fidelity”, “Smooth”, and “Alignment” measure synthesis quality, object identity preservation, motion consistency, and alignment with the text descriptions, respectively. Each metric is rated from 1 (worst) to 4 (best).
}
\vspace{-2mm}

\label{tab:userstudy}
\centering
\resizebox{1.0\linewidth}{!}{
\setlength{\tabcolsep}{24pt}
\begin{tabular}{lcccccccc}
\toprule

& Quality~($\uparrow$)   & Fidelity~($\uparrow$)  & Smooth~($\uparrow$) & Alignment~($\uparrow$)  \\
\midrule
Aether~\cite{aether}  &     1.32 &1.30&1.31&1.34\\

Cosmos(Diff-7B)~\cite{nvidia2025cosmosworldfoundationmodel}        &2.07&2.13&2.05&2.09\\
Cosmos(Diff-14B)~\cite{nvidia2025cosmosworldfoundationmodel}&3.02&2.94&2.98&2.71\\
\method (ours) & \textbf{3.59}                &\textbf{3.63}                  & \textbf{3.65}   &\textbf{3.86}                \\
\bottomrule
\end{tabular}}
\vspace{-7mm}

\end{table*}

\noindent\textbf{Quantitative comparison.} Since there is no method sharing the same setting as ours, we selected two potential competitors for comparison: Cosmos~\cite{nvidia2025cosmosworldfoundationmodel} and Aether~\cite{aether}. As shown in Tab.~\ref{tab:automatic_psnr}, our \method\xspace outperforms all the baselines. Notably, Cosmos~\cite{nvidia2025cosmosworldfoundationmodel} exhibits better generalization ability than Aether~\cite{aether} by explicitly capturing general knowledge of real-world physics and natural behaviors. Besides qualitative results, we provide visualization comparisons in Fig.~\ref{fig:qualitative} as well.
\vspace{-0.5mm}

\noindent\textbf{User study.} In Tab.~\ref{tab:userstudy}, we report the comparison results of human preference rates. We let 20 annotators rate 25 groups of videos, where each group contains the generated video of each method and text description. And we provide detailed regulations to rate the results for scores of 1-4 from four views: ``Quality'', ``Smooth'', ``Fidelity'', ``Alignment''. ``Quality'' counts for whether the result is harmonized without considering fidelity. ``Smooth'' assesses the motion consistency across the video. ``Fidelity'' measures ID preservation and distortions within the video, while we use ``Alignment'' to measure the alignment with the given text descriptions. It can be noted that our model demonstrates significant superiority across all the metrics, especially for ``Alignment'', and ``Smooth''.

\vspace{-3mm}
\section{Conclusion}
\label{sec:conclusion}
\vspace{-3mm}

In conclusion, \method \xspace represents a significant advancement in interactive and realistic world modeling for video generation. Unlike conventional models that are restricted to particular game scenarios or actions, our \method \xspace can capture the complex dynamics of general-world environments and enable free motion control within the simulated world. By formulating world modeling as a joint process of videos and 4D scenes, our \method \xspace ensures coherent world generation and enhances motion and view alignment with the given conditions through part-disentangled motion injection. Experimental results demonstrate our superior performance across diverse scenarios.

\vspace{-0.5mm}
\noindent\textbf{Limitations.} Despite the compelling outcomes, our performance in game scenarios is slightly inferior to realistic ones, likely due to the imbalanced distribution between realistic and game training data. It can be addressed by incorporating more game-scenario datasets in future research.
\paragraph{Acknowledgements.} This work was supported by DAMO Academy via DAMO Academy Research Intern Program.

\bibliographystyle{plain}
\bibliography{camera}

\end{document}